\documentclass{svproc}
\usepackage{url}
\usepackage{graphicx}
\usepackage{caption}
\usepackage{algpseudocode}
\usepackage{algorithm}
\usepackage{amsmath,amsfonts,amssymb}
\usepackage{fixltx2e}
\usepackage{wrapfig}
\usepackage[table,xcdraw]{xcolor}
\usepackage{hyperref}

\begin{document}
\mainmatter              
\title{MNIST Dataset Classification Utilizing k-NN Classifier with Modified Sliding-window Metric}
\titlerunning{k-NN MNIST}  

\author{Divas Grover*, Behrad Toghi*}
\authorrunning{Grover et al.} 
%
\tocauthor{Behrad Toghi, Divas Grover}
\institute{*University of Central Florida, Orlando, FL\\
\email{\{GroverDivas, Toghi\}@knights.ucf.edu}
}

\maketitle              

\begin{abstract}
The MNIST dataset of the handwritten digits is known as one of the commonly used datasets for machine learning and computer vision research. We aim to study a widely applicable classification problem and apply a simple yet efficient K-nearest neighbor classifier with an enhanced heuristic. We evaluate the performance of the K-nearest neighbor classification algorithm on the MNIST dataset where the $L2$ Euclidean distance metric is compared to a modified distance metric which utilizes the sliding window technique in order to avoid performance degradation due to slight spatial misalignments. The accuracy metric and confusion matrices are used as the performance indicators to compare the performance of the baseline algorithm versus the enhanced sliding window method and results show significant improvement using this proposed method.
\end{abstract}
\section{Introduction}

The K-nearest-neighbor (k-NN) classifier is one of the computationally feasible and easy to implement classification methods which sometimes is the very first choice for machine learning projects with an unknown, or not well-known, prior distribution\cite{leif:pablo}. The k-NN algorithm, in fact, stores all the training data and creates a sample library which can be used to classify unlabeled data. During the 70's, k-NN classifier was studied extensively and some of its formal properties were investigated. As an example, authors in [1], demonstrate that for $k=1$ case the k-NN classification error is lower bounded by the twice the Bayes error-rate. Such studies regarding mathematical properties of k-NN led to further research and investigation including new rejection approaches in [2], refinements with respect to Bayes error rate in [3], and distance weighted approaches in [4]. Moreover, soft computing [5] methods and fuzzy methods [6] have also been proposed in the literature.  

A vast literature exists for the classification problem. Among which, one can refer to \cite{yann} by LeCunn et al. where authors have applied different classification algorithms ranging from k-Nearest Neighbor to SVM and Neural Networks. Authors also have used a different kind of pre-processing to increase the accuracy rate. In this work, our main idea is to skip the pre-processing procedure and finding if without any pre-processing we can increase the accuracy over the normal application of k-NN. Thus, we used 10 fold cross-validation to find optimum value of \textit{k} and then applied Sliding Window technique which is commonly used in Machine Vision to detect different objects in a frame. Utilizing the sliding window technique, performance degradation, due to a minor spatial displacement between test and training, has been avoided. Accuracy, confidence interval, and confusion matrices are used to evaluate the model’s prediction performance.

The rest of the paper is organized as follows, in section 2 we'll see the basic outline of k-NN and how it's implemented, in section 3 we have defined our process of cross-validation and our experiments with simple k-NN which is followed by section 4, explaining the Sliding Window and experiments ran on it. Finally, we compare the above methods in Section 5 and conclude the experiments in Section 6.                                
\section{Baseline k-NN Implementation}

The k-NN algorithm relies on voting among the k nearest neighbors of a data point based on a defined distance metric. The distance metric is chosen considering the application and the problem nature and it can be chosen from any of the well-known metrics, e.g., Euclidean, Manhattan, cosine, and Mahalanobis, or defined specifically for the desired application. In our case, we utilize a Euclidean distance metrics which is defined as follows:
\begin{equation}
\vec{D}^{n\times m}(\vec{x^n}, \vec{y^m}) = \left[\sum_{i = 0}^{28 \times 28 - 1} {(\vec{{x_{i}^n}} - \vec{y_{i}^m})}^2\right]^{\frac{1}{2}}
\end{equation}
where,
\begin{displaymath}
 \;\vec{x^n};  n \in \{0,59999\} \text{ is the }\;Train Set
\end{displaymath}
\begin{displaymath}
\vec{y^m};  m \in \{0,9999\} \text{ is the }\;Test Set
\end{displaymath}

\noindent
\begin{wrapfigure}{R}{0.4\textwidth}
\centering
\caption{Sample digits from the testset}
\label{fig:viz}
\includegraphics[width=0.5\textwidth]{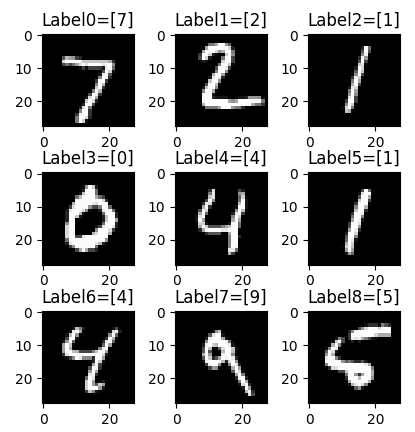}
\end{wrapfigure}

As it is mentioned before, we are using a refined version of the MNIST database of handwritten digits which consists of $60,000$ labeled training images as well as $10,000$ labeled test images. Every data-point, i.e., a test or training example is a square image of a handwritten digit which has been size-normalized and centered in a
fixed-size $28 \times 28$ pixel image\cite{yann:cortes}. Figure \ref{fig:viz} show an example visualization of the first $50$ digits from the above-mentioned test set. Similar digits can have various shapes and orientation in the dataset which means, in the extreme case, distance, i.e., Euclidean distance, between two alike digits can be greater than that of two non-alike digits. This virtue adds more complexity to the classification process and may cause performance degradation of the model. Thus, pre-processing the data can help to mitigate such classification errors and consequently improve the model’s performance.

\begin{figure}
\centering

\caption{Average Euclidean distance between a sample “one” digit and other training images}
\label{avg:dist}
\includegraphics[height = 6cm]{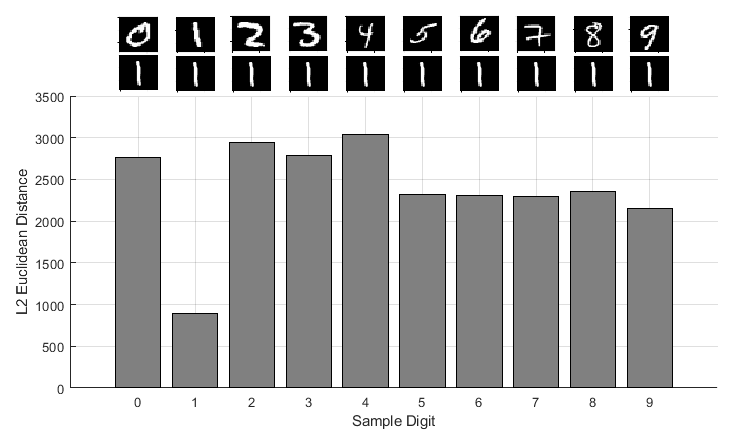}
\end{figure}

As it is demonstrated in Algo 1, we use a 2-dimensional matrix to store all distance pairs between test and training data points, i.e.$[60,000 \times 10,000]$,  distances. Once this distance matrix is derived, there is no need to calculate distances on every iteration, for example for different k-values. This significantly increases the time-efficiency and decreases the run times. The $i^{th}$ row of the distance matrix stores the distance between the $i^{th}$ test image and all 60,000 training images. Thus, for every test image, the $i^{th}$ row of the matrix is sorted and k lowest distances are extracted alongside their corresponding indices. These indices can be used for comparison to the original saved labels in order to evaluate the classification accuracy. In order to attain more intuition on the problem, we calculate the average distance between a sample digit and all other digits in the data set; results are shown in Figure \ref{avg:dist}.

As it is shown in Figure \ref{avg:dist}, the distance between two alike digits is obviously less than that of two different digits. However, in some cases where digits are visually similar, e.g., eight and zero, the distance between two not-alike digits can be also small. The next step, after getting the nearest neighbors to a test image, would be reporting the evaluation metrics. We chose accuracy metric, confidence interval, and confusion matrix as our evaluation metrics. Accuracy can be obtained from the following equation:
\begin{equation}
ACC = \frac{TP + TN}{P + N}
\end{equation}
where $TP$ and $TN$ denote the number of True Positive and True Negative instances, respectively. $P$ and $N$ are total positive and total negative samples respectively.

\section{k-fold Cross Validation}

\begin{figure}[h!]
\centering
\caption{10-fold cross validation procedure and deriving optimal k-value }
\label{crs:fld}

\includegraphics[width=0.8\textwidth]{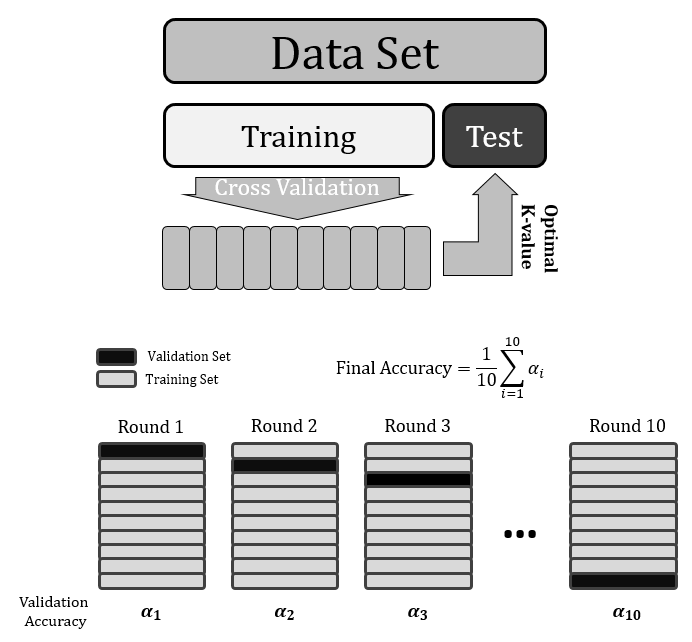}
\end{figure}

\begin{figure}[t]
\centering
\caption{Obtained accuracy for k values in range $[0, ...,10]$}
\label{crs:val}
\includegraphics[height = 5cm]{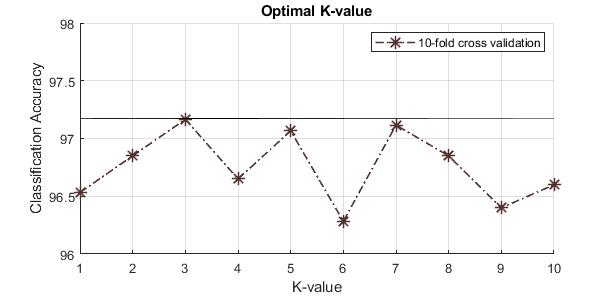}
\end{figure}

As it is mentioned in the previous sections, the choice of the k-value can impact the classifier’s performance. Hence, we conduct a k-fold cross-validation procedure to evaluate the obtained accuracies for k-values in the range [0,10] and consequently choose the optimal k-value \cite{pedregosa11a}. The cross-validation procedure is illustrated in Figure \ref{crs:fld}. The cross-validation is conducted over the training set by dividing it into 10 slices, each containing 6,000 images. Then the total accuracy for every k-value is derived by averaging the accuracy values for each fold.  Table \ref{tab:cross_val} tabulates the results for cross-validation procedure over the desired k-values also in Figure \ref{crs:val} we can see how accuracy changes by varying the $k$ value. Results demonstrate that the optimal k-value for the dataset under test should be. We use this value for the rest of our study.
\begin{algorithm}[h!]
\caption{Pseudo code of Algorithm}\label{erfsffgfg}
\begin{algorithmic}[1]
\Function{<MAIN>}{}
\State \texttt{train, test = LOAD\_DATASETS()}
\State \texttt{change train and test to np.array() class}
\State \texttt{cross\_val = 10x10 null array to store accuracies}\\
\For{i in range (1,10)}
    \State \texttt{new\_train, val\_set, ind =  CROSS\_VALIDATION(train, i)}
    \State \texttt{Extract labels from first column of both new\_train and val\_set}
    \State\texttt{and store them as lbl\_trn and lbl\_val}
    \\
    \State \texttt{dist\_matrix = EUCLIDIAN\_DISTANCE(new\_train, val\_set)}
    \State{\textit{dist\_matrix is an array of 6,000 by 54,000, having distance of each validation example}}
    \State{\textit{to every training example}}
    \\
    \For{k in range (1,11)}
        \State\texttt{neigh = NEIGHBOR(dist\_matrix, k)}
        \State\textit{neigh is the a matrix of 6,000 by k which contains indices of}
        \State\textit{k Nearest Neighbors in Training Set.}
        \\
        \State\texttt{prd = GET\_LABEL(neigh, lbl\_trn)}
        \State\textit{prd is a vector of 6,000 having predicted labels based on the indices stored in neigh}
        \\
        \State\texttt{crr = no. of correct labels; by comparing prd and lbl\_val}
        \State\texttt{accuracy = (crr divided by length of lbl\_val)x100}
        
        \State\texttt{add accuracy of ith validation set and for present value}
        \State\texttt{of k to cross\_val(i)(k-1)}
      
    \EndFor
\EndFor
\State save cross\_val as csv file
\EndFunction
\end{algorithmic}
\end{algorithm}
\begin{table}[]
\caption{Performance Analysis k=3}
\centering
\begin{tabular}{|l|l|}
\hline
\rowcolor[HTML]{FFFFFF} 
Accuracy(\%)             & 97.17                 \\ \hline
Standard Deviation  & 0.001658              \\ \hline
\rowcolor[HTML]{FFFFFF} 
Confidence Interval & {[}0.97004,0.97335{]} \\ \hline
\end{tabular}
\label{tab:perfana}
\end{table}

\begin{figure}[h!]
\centering

\caption{Confusion Matrix k=3}
\label{conf:1}
\includegraphics[height = 8cm]{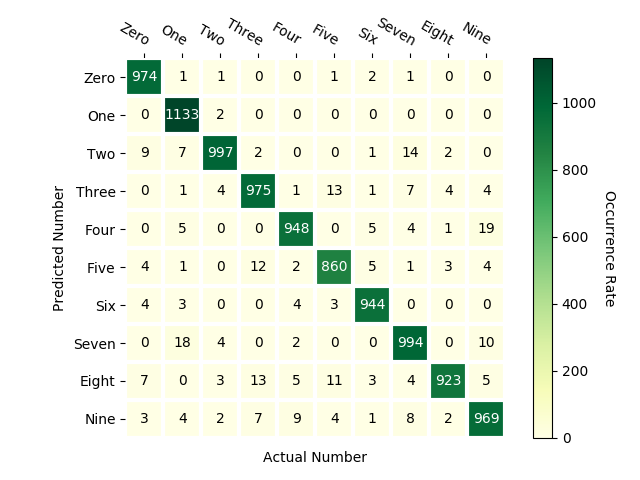}
\end{figure}

Figure \ref{conf:1} illustrates the confusion matrix for the classification model with $k=3$. Accuracy and confidence interval results are also reported in Table \ref{tab:perfana}. The standard deviation of a Binomial distribution can be estimated as
\begin{equation}
\hat{\sigma} = \sqrt{\frac{\hat{p}\times{(1-\hat{p})}}{n}}
\end{equation}
where $\hat{p}$ is calculated accuracy. Considering $95\%$ Confidence Interval, the z-score will $1.96$ which leads to confidence interval as shown below.
\begin{equation}
\text{c.i.:} = \mu\pm1.96\times\hat{\sigma}
\end{equation}

A confusion matrix is often used as a graphic to visualize a classifier's performance. It's a quantitative plot of Actual Classes vs. the Predicted Classes. The actual classes are horizontal and predicted are vertical in Figure \ref{conf:1}. We can clearly see that diagonally we have bigger counts because the classifier maps actual class to correct predicted class most of the times and in other cells, we can see that how many times a certain actual class is misclassified.  
 
\begin{table}[h!]
\centering
\caption{10-Fold Cross Validation}
\label{my-label}
\begin{tabular}{|l|l|l|l|l|l|}
\hline
\rowcolor[HTML]{C0C0C0} 
k-value     & 1     & 2     & 3     & 4     & 5     \\ \hline
Accuracy (\%) & 96.53 & 96.84 & 97.17 & 96.64 & 97.06 \\ \hline
\rowcolor[HTML]{C0C0C0} 
k-value     & 6     & 7     & 8     & 9     & 10    \\ \hline
Accuracy (\%) & 96.28 & 97.11 & 96.84 & 96.39 & 96.59 \\ \hline
\end{tabular}
\label{tab:cross_val}
\end{table}

\section{Sliding Window L2 Metric}

\begin{figure}[t]
\centering
\caption{Sliding Window Illustration}
\label{sld:win}
\includegraphics[width = \textwidth, height = 6cm]{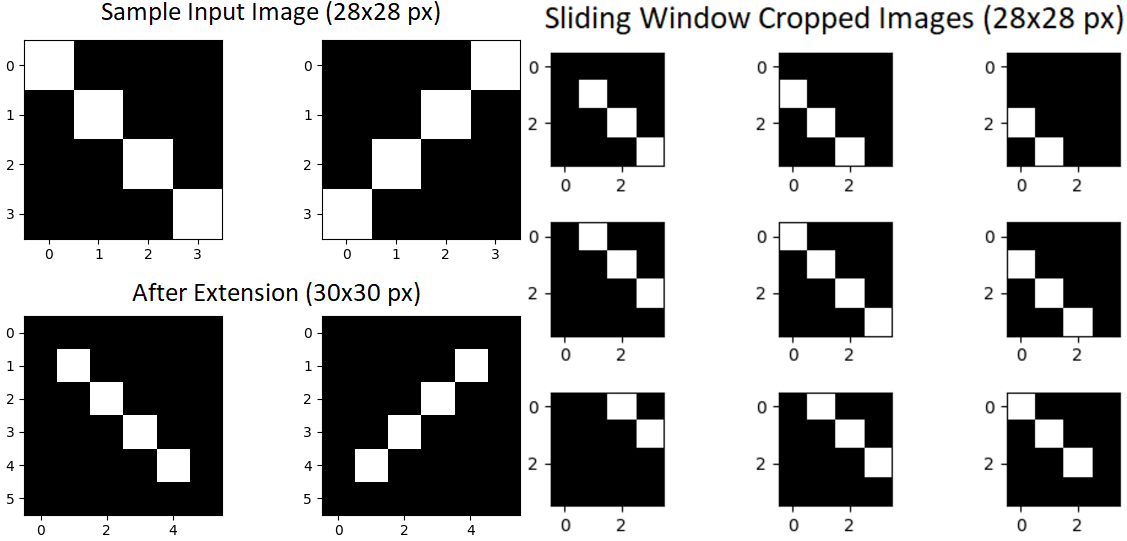}
\end{figure}

\begin{figure}
\centering
\caption{Confusion Matrix k=3 Sliding Window}
\label{conf:2}
\includegraphics[height = 8cm]{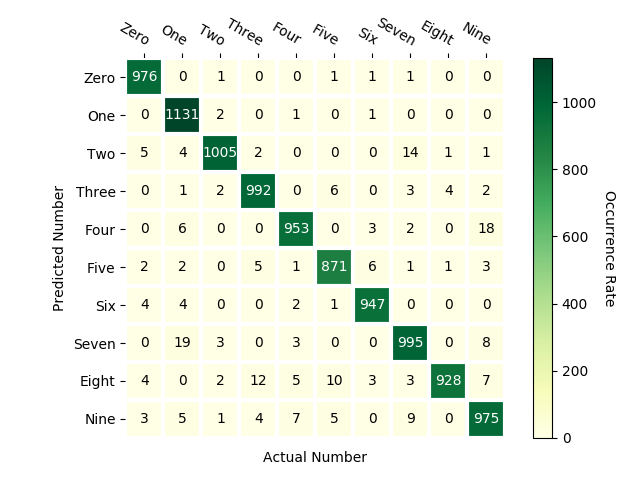}
\end{figure}

The idea in this section is to mitigate the false distance measurements due to the small spatial translations in the image under test or the training images. As mentioned before, every example image is a square $28\times28$ The idea in this section is to mitigate the false distance measurements due to the small spatial translations in the image under test or the training images. We pad all the training examples by $0$s and they become squares of $30\times30$ then, every extended image is cropped by sliding a square $28\times28$ window over 9 possible positions. Hence, every image produces 9 versions among which one is the original image itself\cite{glumov:sergeyev}. For the sake of simplicity, we use a simple black and white diagonal input image to show the sliding window process in Figure \ref{sld:win}. Accuracy and confidence interval are reported in Table \ref{tab:perf_ana_sliding}. After applying the sliding window on the dataset we can also visualize the modified classifier performance through confusion matrix shown in Figure \ref{conf:2}.
A modified distance metric $\widetilde{D}^{(n\times m)}$, can be introduced to summarize the sliding window method 

\begin{equation}
\vec{\widetilde{D}^{(n\times m)}}{(\vec{x^n}, \vec{y^m})} = min\left[\vec{\widetilde{D}^{(n\times i \times m)}}{(\vec{x^{n,i}}, \vec{y^m})}\right]
\end{equation}
where $i = 1, ..., 9$.
\section{Classifier Accuracy}

We conduct hypotheses evaluation in order to compare the performance of our two methods, i.e., baseline k-NN and sliding window k-NN. We define the null and alternative hypotheses as follow
\begin{displaymath}
\begin{cases}
\mathcal{H}_0: p_{baseline} = p_{sliding}\\
\mathcal{H}_a: p_{baseline} < p_{sliding}
\end{cases}
\end{displaymath}

The significance level, $\alpha$ is defined as the test’s probability of incorrectly rejecting the null hypothesis:
\begin{equation}
\alpha = P_{r}(|\hat{d}>z_{N}\sigma|)
\end{equation}
\begin{table}
\centering
\caption{Performance Analysis k=3 Sliding Window}
\begin{tabular}{|l|l|}
\hline
\rowcolor[HTML]{FFFFFF} 
Accuracy (\%)             & 97.73\%                 \\ \hline
Standard Deviation  & 0.001489              \\ \hline
\rowcolor[HTML]{FFFFFF} 
Confidence Interval (c.i.) & {[}0.97581,0.97879{]} \\ \hline
\end{tabular}
\label{tab:perf_ana_sliding}
\end{table}
\\or in other words $\alpha = 1- N$, where N is the confidence interval of 95\% in our case. We then reject the hypothesis if the test statistic $z = \frac{|\hat{d}|}{\sigma_{\hat{d}}}$ is greater than $z_{N} = 1.96$ (for $n = 95$). Assuming the normal distribution, $\hat{d} = 0.0022$ or $0.2$ which can be considered rule of thumb difference criteria. In our case we have $d = 0.9773 - 0.9717 = 0.0056$. Hence $d > \hat{d}$ which means the null hypothesis is rejected and there is evidence for the alternative hypothesis.
\section{Concluding Remarks}
We implemented a k-NN classifier with Euclidean distance and an enhanced distance metric utilizing the sliding window method. Our results show a significant improvement by employing the enhanced metric. Our algorithm relies on removing the spacial shifts from the test and training set data in a way that avoids misclassification resulting from co-labeled but shifted images. We present our results in terms of performance indicators such as accuracy metric and confusion matrices.
\section{Source Code}
The source code of our implementation is publicly available on GitHub\footnote { \href{https://github.com/BehradToghi/kNN_SWin}{https://github.com/BehradToghi/kNN\_SWin} }. The current implementation is compatible with the refined MNIST dataset, also available online \cite{mnist}.

%
%

\end{document}